\documentclass[letterpaper, 10 pt, conference]{ieeeconf}  

\IEEEoverridecommandlockouts                             

\usepackage{graphicx} 
\usepackage{amsmath}
\usepackage{bm}
\usepackage{xcolor}
\usepackage{lipsum}
\usepackage{amssymb}  

\usepackage[style=ieee]{biblatex}

\DeclareSourcemap{
  \maps{
    \map{
      \pertype{article}
      \step[fieldset=url, null]
      \step[fieldset=doi, null]
      \step[fieldset=issn, null]
      \step[fieldset=isbn, null]
      \step[fieldset=note, null]
      \step[fieldset=editor, null]
      \step[fieldset=urldate, null]
      \step[fieldset=file, null]
    }
  }
}
\DeclareSourcemap{
  \maps{
    \map{
      \pertype{inproceedings}
      \step[fieldset=url, null]
      \step[fieldset=doi, null]
      \step[fieldset=issn, null]
      \step[fieldset=isbn, null]
      \step[fieldset=note, null]
      \step[fieldset=editor, null]
      \step[fieldset=urldate, null]
      \step[fieldset=file, null]
    }
  }
}
\DeclareSourcemap{
  \maps{
    \map{
      \pertype{incollection}
      \step[fieldset=url, null]
      \step[fieldset=doi, null]
      \step[fieldset=issn, null]
      \step[fieldset=isbn, null]
      \step[fieldset=note, null]
      \step[fieldset=editor, null]
      \step[fieldset=urldate, null]
      \step[fieldset=file, null]
    }
  }
}

\title{\LARGE \bf
Thruster-Assisted Incline Walking 
}
\author{Kaushik Venkatesh Krishnamurthy$^{1}$, Chenghao Wang$^{1}$, Shreyansh Pitroda$^{1}$, Adarsh Salagame$^{1}$, Eric Sihite$^{2}$, \\ Reza Nemovi$^{2}$, Alireza Ramezani$^{1*}$,  Morteza Gharib$^{2}$ 
\thanks{$^{1}$The author is with Department of Electrical and Computer Engineering,
        Northeastern University, Boston, MA, USA  {\tt\small venkateshkrishnamu.k, wang.chengh, pitroda.s, a.salagame@northeastern.edu}}%
\thanks{$^{2}$The author is with the Department of Aerospace Engineering, California Institute of Technology, Pasadena, CA, USA {\tt\small rnemovi, esihite, mgharib@caltech.edu}%
}
\thanks{$^{*}$Corresponding author {\tt\small a.ramezani@northeastern.edu}}}

\begin{document}

\maketitle
\thispagestyle{empty}
\pagestyle{empty}

\begin{abstract}
In this study, our aim is to evaluate the effectiveness of thruster-assisted steep slope walking for the Husky Carbon, a quadrupedal robot equipped with custom-designed actuators and plural electric ducted fans, through simulation prior to conducting experimental trials. Thruster-assisted steep slope walking draws inspiration from wing-assisted incline running (WAIR) observed in birds, and intriguingly incorporates posture manipulation and thrust vectoring, a locomotion technique not previously explored in the animal kingdom. Our approach involves developing a reduced-order model of the Husky robot, followed by the application of an optimization-based controller utilizing collocation methods and dynamics interpolation to determine control actions. Through simulation testing, we demonstrate the feasibility of hardware implementation of our controller.
\end{abstract}
\maketitle

\section{Introduction}
\label{sec:intro}

WAIR (Wing Assisted Inclined Running) was reported in \cite{dial_wing-assisted_2003,tobalske_aerodynamics_2007,peterson_experimental_2011} upon the observation that certain birds possess the ability to utilize their wings to flap, generating aerodynamic forces that augment tractive forces. However, WAIR faces the challenge of balancing the need for sufficient tractive forces, influenced by aerodynamics, while ensuring that Ground Reaction Forces (GRF) remain within friction cone constraints. Birds demonstrate remarkable dexterity in running across a wide range of slopes, both rough and smooth, exhibiting high locomotion plasticity. This natural adaptation allows birds to traverse expansive terrains and expand their habitat.

The integration of posture manipulation and thrust vectoring, as seen in WAIR exhibited by birds, presents complex locomotion challenges worthy of study. Contemporary legged robots can navigate exceedingly rough terrain \cite{fankhauser_robust_2018,griffin_footstep_2019}, achieve running speeds \cite{hyun_high_2014, shin_design_2022}, execute high jumps \cite{sato_vertical_2021, dai_whole-body_2014}, and even engage in parkour \cite{hoeller_anymal_2024, cheng_extreme_2023}. Similarly, multi-rotor systems demonstrate agile maneuvers, sometimes surpassing birds in agility. However, an aspect of multi-modal systems akin to birds \cite{dirckx_optimal_2023,song_policy_2022}, which remains less explored, is the collaborative operation of aerodynamically proficient components like wings alongside legs—specifically, through posture manipulation and thrust vectoring—to overcome locomotion challenges. Robots equipped with capabilities akin to WAIR-performing birds can surmount significant locomotion hurdles, such as traversing steep slopes, navigating narrow paths, hurdling over large obstacles, and more. Slope walking can also become a task that becomes harder for robots as motor capabilities are far from living animals. Strategies for slope walking are also different as explained in \cite{gehring_dynamic_2015} and were able to perform dynamic trotting on slopes with a careful foothold selection strategy.

\begin{figure}[t]
    \centering
    \includegraphics[width = \linewidth]{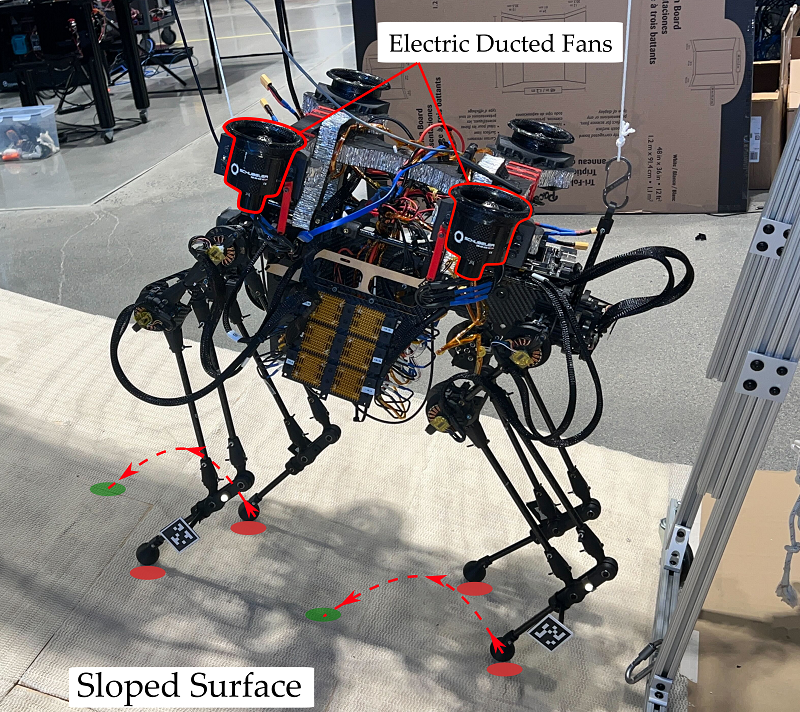}
    \caption{Capturing Northeastern Husky Carbon, a quadrupedal robot with four ducted fans affixed to its torso, undergoing WAIR locomotion tests inspired by avian biomechanics, as it navigates a ramp.}
    \label{fig:cover-image}
\end{figure}

\begin{figure*}[t]
    \centering
    \includegraphics[width=0.8\linewidth]{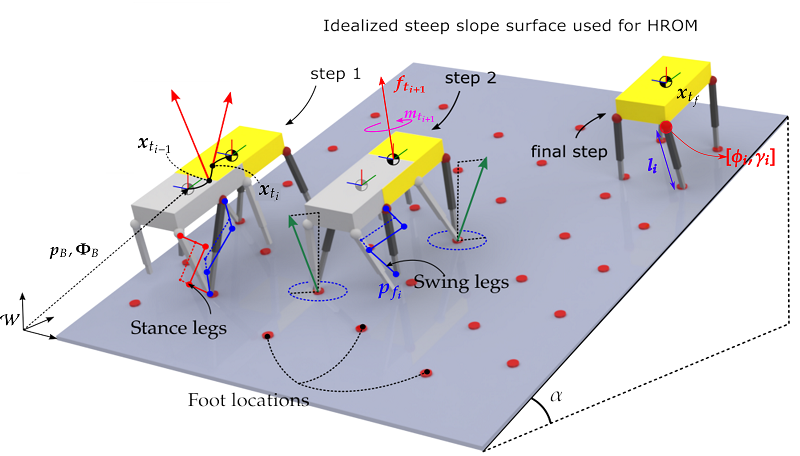}
    \caption{Depicts the parameters of the reduced-order model (ROM) for the Husky, which are employed to govern the equations of motion as detailed in Section~\ref{sec:mdl}.}
    \label{fig:wair}
\end{figure*}

The Northeastern University Husky Carbon \cite{ramezani_generative_2021,wang_legged_2023,liang_rough-terrain_2021,sihite_unilateral_2021,sihite_optimization-free_2021, venkatesh_krishnamurthy_towards_2023,salagame_quadrupedal_2023,sihite_optimization-free_2021,sihite_dynamic_2023,liang_rough-terrain_2021,sihite_efficient_2022-1} has been conceived and developed at Northeastern University in Boston. This quadrupedal robot features custom-designed actuators and body structure, equipped with four legs and four electric ducted fans (EDFs). Each leg boasts three degrees of freedom, actuated by three custom-designed permanent DC actuators with harmonic drives. These actuators are powered by high-power ELMO amplifiers, enabling torque control through current regulation in the DC actuator windings.

The hip-sagittal (HS) joint works in tandem with the knee (K) joint to maneuver the leg in the hip sagittal plane. In the interest of this research, the hip frontal (HF) joints play a very important role by being able to control the position of the foot in the frontal plane of the robot. With three motors to control the above three joints, all the 12 joints on the robot are actuated by T-motor Antigravity 4006 brushless motors, with the motor output transmitted through a Harmonic drive. The Harmonic drives are chosen for their precise transmission, low backlash, and back-drivability. The motor and gearbox housings along were embedded in the housing during the printing process making the robot's legs significantly lightweight.

The overarching objective of the Husky Carbon design is multifaceted. Firstly, we aim to explore multi-modal locomotion through appendage repurposing \cite{sihite_multi-modal_2023,mandralis_minimum_2023, sihite_demonstrating_2023, sihite_dynamic_2023}. Secondly, our goal is to push the boundaries of locomotion beyond those faced by standard legged systems \cite{buss_preliminary_2014,park_finite-state_2013}, achieved through the integration of posture manipulation and thrust vectoring. Traditional legged robots are constrained as they can only manipulate contact forces via posture manipulation, limiting their operational capabilities (e.g., traversing steep surfaces becomes challenging). Although recent studies have reported successful locomotion on vertical surfaces \cite{hu2022magnetic}, these works are confined to metal surfaces, relying on electromagnetic principles to establish bonds at contact points.

The paper introduces a constrained optimal controller based on collocation method designed to determine the optimal friction cone-admissible thruster inputs, represented as a wrench about the COM. This controller utilizes a Reduced-Order Model (ROM) of the Husky, considering a maximum of two feet in contact with the ground and solves for the ground forces using Lagrange multipliers. The structure of the paper is organized as follows: First, we detail the modeling strategy employed. Then, we explain the design of the controller. Finally, we present Matlab simulation results depicting WAIR performance over slopes and outline future research directions in our concluding remarks.

\begin{figure*}
    \centering   \includegraphics[width=\linewidth]{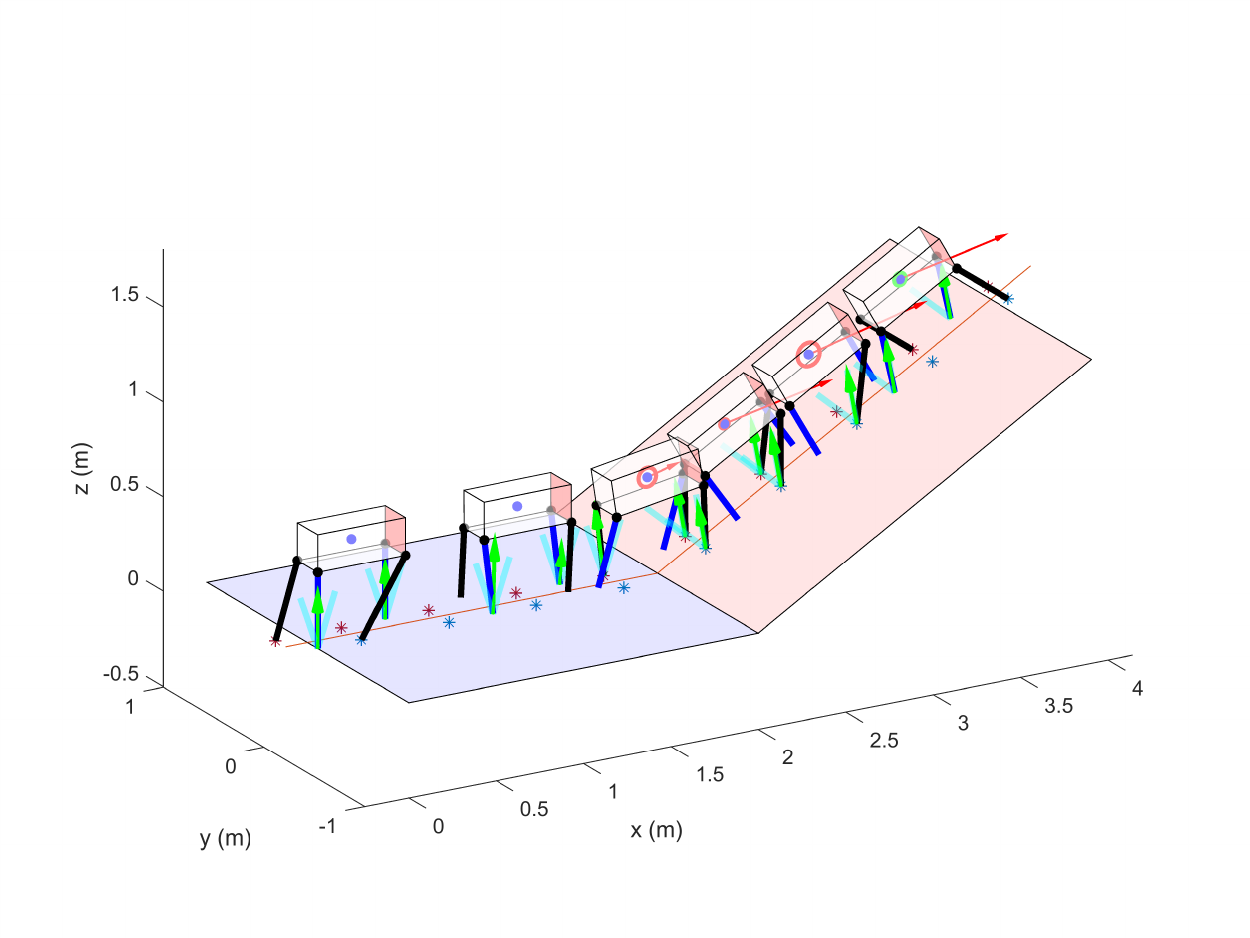}
    \caption{Illustrates snapshots of the WAIR simulation performed in Matlab on a slope of 30 degrees. The figure also shows the foot locations, Center of mass trajectory, optimal thrust wrench inputs (red), scaled ground reaction (green) forces and the positions of the legs. It can be seen that the ground reaction force stays within the friction cone}
    \label{fig: snapshots}
\end{figure*}





\section{Modelling}
\label{sec:mdl}

The Husky ROM and its parameters are shown in Fig.~\ref{fig:wair}. As shown in this figure the following simplifications are assumed: The body is assumed to a single rigid body shaped as a cuboid of homogeneous mass with the hips located at the lower corners of the cuboid. By assuming sufficient torque authority in the legs, the legs are assumed to be massless with a simplified spherical join in the hip followed by a prismatic joint for the length of the joint. The thruster forces are approximated as a wrench in acting on the center of mass (hereby referred to as 'thrust wrench').

The equations of motion of the HROM can be derived using the energy-based Euler-Lagrange dynamics formulation. As illustrated in Fig.~\ref{fig:wair}, the positions of the leg ends are defined as functions of the spherical joint primitives, namely $\phi$ and $\gamma$, along with the length of the leg $l$. The pose of the body can be defined using $\bm{p}_B \in \mathbb{R}^3$, and Z-Y-X Euler angles $\bm{\Phi}_B$. The rotation matrix can also then be defined from the Euler angles as $R_B$. The generalized coordinates of the robot body can then be defined as follows:
\begin{equation}
\bm{q} = [\bm{p}_B^\top ,\bm{\Phi}_B^\top ]^\top 
\label{eq: body states}
\end{equation}
And, the leg states of the robot can be defined as,
\begin{equation}
\begin{aligned}
\bm{q}_L &= [\dots, \phi_i,\gamma_i,l_i, \dots]^\top \\
i &\in \mathcal{F},
\label{eq: leg states}
\end{aligned}
\end{equation}
where $\mathcal{F} =\left\{FR, HR , FL , HL\right\}$ represents the respective legs and thrusters. The position of the foot can then be determined using the forward kinematics equations shown:
\begin{equation}
\begin{aligned}
    \bm{p}_{f i} &= \bm{p}_{B} + R_{B} \bm{l}_{h i}^{B} + R_{B} \bm{l}_{f i}^{B} \\
    \bm{l}_{f i}^{B} &= R_{y}\left(\phi_{i}\right) R_{x}\left(\gamma_{i}\right)
    \begin{bmatrix} 
    0, & 0, & -l_{i}
\end{bmatrix}^\top
\label{eq:foot_pos}
\end{aligned}
\end{equation}
The positions of the thrusters are defined as $\bm{p}_{ti}$ with respect to the body as follows, 
\begin{equation}
    \bm{p}_{ti} = \bm{p}_B + R_B\bm{l}_{t,i}^B
\end{equation}
The superscript $B$ denotes a vector defined in 
the body frame, while the rotation matrix $\bm{R_B}$ represents the rotation of a vector from the body frame to the inertial frame. As the legs are considered massless, the kinetic and potential energies of the HROM can be calculated using the equations provided below:
\begin{equation}
\begin{aligned}
    \mathcal{K} &= \left( \frac{1}{2} \dot{\bm{p}}_B m_B \dot{\bm{p}}_B^\top +\bm{\omega}_B^B I_B \bm{\omega}^{B\top}_B \right) \\
     \mathcal{V} &= -m_B \bm{p}_B^\top \bm{g} \\
    \mathcal{L} &= \mathcal{K}-\mathcal{V},
\label{eq:LKV}
\end{aligned}
\end{equation}
where $\bm{\omega}_B^B$ represents the body angular velocity in the body frame, and $\bm{g}$ denotes the gravitational acceleration vector. The angular velocity of the body can be found as a function of the rate of change of Euler angles using the Euler rate matrix $E(\bm{\Phi})$,
\begin{equation}
\bm{\omega}_B^B = E(\bm{\Phi_B}) \dot{\bm{\Phi}}_B
\end{equation}
 The dynamic equation of motion can be derived using the Euler-Lagrangian method as follows:
\begin{equation}
    \textstyle \frac{d}{dt}\left(\frac{\partial{\mathcal{L}}}{\partial{\bm{v}}}\right)- \frac{\partial \mathcal{L}}{\partial{\bm{q}}} = \bm{\Gamma},
\label{eq:euler-lagrangian}
\end{equation}
where $\bm{\Gamma}$ is the sum of all generalized and constraint torques and forces respectively. The dynamic system accelerations can then be solved to obtain the into the following standard form:
\begin{equation}
    \bm M\dot{\bm v} + \bm h = \bm u_e + \sum_{i \in \mathcal{F}}\bm J_{i,l}(\bm q)^{\top} \bm \lambda_i,
    \label{eq: eom}
\end{equation}
where $\bm M $ is the mass/inertia matrix, $\bm h$ contains the Coriolis and gravitational vectors, and $\bm v$ contains the generalized velocities. The term $\bm{u}_e~\in~\mathbb{R}^6$ represents the external thrust wrench acting on the COM of the rigid body of the HROM, i.e., $\bm u_e~=~[\bm{f_t}^{\top},\bm{m_t}^{\top}]^{\top}$, where $\bm{f_t}$ and $\bm{m_t}$ are the forces and moments forming the  thrust wrench . 

Then, $\bm J_{i,l}(\bm q) ~=~\frac{\partial{\dot{\bm{p}}_{f,i}}}{\partial{\bm{v}}}~\in~\mathbb{R}^{3 \times n}$ represents the Jacobian of the contact point of the $i^{\mathrm{th}}$ leg end with Cartesian coordinates $\bm{p}_{f,i}$, and $\bm \lambda_i \in \mathbb{R}^3$ is the Lagrange multiplier. The equations of motion can then be written in the following form:
 \begin{equation}
\begin{gathered}
    \dot{\bm{x}} = \bm{f}(\bm{x},\bm{u}), \\
    \bm{x} = [\bm q_d^\top ,\bm{v}_d^\top ]^\top\\
    \bm{u} = [\bm{u}_e^\top, \bm{u}_L^\top]^\top\\
    \end{gathered}
\label{eq:eom}
\end{equation}
where $\bm q_d = [\bm q^\top, \bm q_L^\top]^\top$, $\bm{v}_d = \left[\bm{v}^\top, \bm{\dot{\bm q}}_L^\top\right]^\top$, and $\bm{x}$ is obtained by combining both the dynamic and massless leg states and their derivatives to form the full system states. The constraint equations for the Lagrange multipliers are then defined as:
\begin{equation}
\begin{gathered}
        \bm J_{i,l}(\bm q_d) \dot{\bm q}_d = 0 \\
    \bm J_{i,l}(\bm q_d) \ddot{\bm q}_d + \left( \frac{\bm J_{i,l}(\bm q_d)\dot {\bm q}_d}{\partial \bm{q_d}} \right) \dot{\bm q}_d = 0, \quad \forall i \in \mathcal{F}
\end{gathered}
\label{eq: lagrange multiplier constraint}
\end{equation} 
The Lagrange multipliers $\bm{\lambda}_i$ can then be solved for using the system of equations formed in Eq.~\ref{eq: eom} and Eqs.~\ref{eq: lagrange multiplier constraint}. The constraint equations arises from the assumption that contact is rigid and there is no slippage. The wrench due to the GRF is then calculated as:
\begin{equation}
    \bm w = \bm{J_l}^{\top} \bm{\lambda}
\end{equation}
The no-slippage condition is ensured since in the subsequent step, outlined in Section~\ref{sec:ctrl}, we obtain the control actions that maintain the states within the constrained-admissible set.

\section{Collocation-based controller}
\label{sec:ctrl}

Our objective is to leverage collaboration between foot placement (posture manipulation) and thruster wrench (thrust vectoring) to maintain contact forces within the friction cone. To achieve this goal, we employ an optimization-based method utilizing collocation.

To solve for the controls problem, we find the optimal wrench about the COM using the following cost function, 
\begin{equation}
    \centering
    \begin{aligned}
    J =& \frac{1}{2}\lVert \bm f_t \rVert ^2 \bm w_1 + \frac{1}{2}\lVert \bm m_t \rVert ^2 \bm w_2
    \end{aligned}
\end{equation}
where $\bm x_e$ represents an error term that calculates the discrepancy of the body pose from the Body Euler angles. $\bm w_1, \bm w_2 \in \mathbb{R}^3$ are scalar weights. The cost is determined by the model outlined in Eq.~\ref{eq: eom}. Through temporal discretization of the dynamics, we obtain equations of the following form:
\begin{equation}
\dot{\bm{x}}_i=\bm{f}_i(\bm{x}_i,\bm{u}_i), \quad i=1, \ldots, n, \quad 0 \leq t_i \leq t_f,
\label{eq:discrete-hrom-model} 
\end{equation}
where the vector $\bm x_i$ contains the values of the state vector at the $i^{\mathrm{th}}$ discrete time step, $u_i$ contains the values of the thrust  wrench at the $i^{\mathrm{th}}$ sample time, and $\bm f_i$ denotes the governing dynamics at the $i^{\mathrm{th}}$ discrete time step. 

The discrete values, $\bm x_i$ and $\bm u_i$, are stacked in the vectors $\bm X = \left[\bm{x}^\top_1(t_1), \ldots, \bm{x}_k^\top(t_k)\right]^\top $ and $\bm U =~\left[\bm{u}^\top_1(t_1), \ldots, \bm{u}^\top_k(t_k)\right]^\top $. The boundary conditions for the time period are given by:
\begin{equation}
    r_i\left(\bm{x}(0), \bm{x}\left(t_f\right), t_f\right)=0, \quad i=1, \ldots, 2 n
\end{equation}
The input thrust wrench consists of 6 entries, thus requiring consideration of atleast 6 inequality constraints to ensure that the thruster forces remain within the admissible set defined by the friction cone constraint. Other constraints include maintaining a minimum normal force and also bounding the magnitude of the input. Put together, these constraints take the form,
\begin{equation}
    \bm{g}_i(\bm{x}(t_i), \bm{u}(t_i), t_i) \geq 0, \quad i=1,\dots,m \quad 0 \leq t_i \leq t_f
\end{equation}
To approximate the nonlinear dynamics from HROM, we employ a method based on polynomial interpolations. This method extremely simplifies the computation efforts. Consider the $n$ time intervals, as defined previously and given by 
\begin{equation}
0=t_1<t_2<\ldots<t_n=t_f
\end{equation}
The optimal solutions for the WAIR maneuver are determined for a fixed $t_f$, but it can also be included as a parameter in the optimization problem, thereby enabling the optimizer to determine the walking speed as well. The decision parameter vector can be denoted as $\mathcal{Y}$, where:
\begin{equation}
    \mathcal{Y} = \left[ \bm X ; \bm U ; t_f\right] 
\end{equation}
To find the decision parameters $\bm X$ and $\bm U$, we use the Matlab's nonlinear optimization function \textit{fmincon}. To solve this rapidly, we use an interpolation method to find $x_i$ and $u_i$. The input and is formed as a linear interpolation function between $\bm u_i$ and $\bm u_{i+1} $ for $t_i \leq t \leq t_{i+1}$, i.e $\Tilde{\bm u}$
\begin{equation}
\tilde{\bm{u}}= \bm{u}_i\left(t_i\right)+\frac{t-t_i}{t_{i+1}-t_i}\left( \bm{u}_{i+1}\left(t_{i+1}\right)-\bm{u}_i\left(t_i\right)\right)
\end{equation}
The states $\bm x_i$ and $\bm x_{i+1}$ also need to be interpolated and a cubic interpolation that is continuously differentiable with $\Tilde{\dot{\bm x}} = \bm{f}(\bm x(s), \bm u(s), s)$ at $s = t_i$ and $s = t_{i+1}$. Then, $\tilde{\bm{x}}$, is found as follows, 
\begin{equation}
    \begin{aligned}
\tilde{\bm{x}}(t) &=\sum_{k=0}^3 c_k^j\left(\frac{t-t_j}{h_j}\right)^k, \quad t_j \leq t<t_{j+1}, \\
c_0^j &=\bm{x}\left(t_j\right), \\
c_1^j &=h_j \bm{f}_j, \\
c_2^j &=-3 \bm{x}\left(t_j\right)-2 h_j \bm{f}_j+3 \bm{x}\left(t_{j+1}\right)-h_j \bm{f}_{j+1}, \\
c_3^j &=2 \bm{x}\left(t_j\right)+h_j \bm{f}_j \bm{x}\left(t_{j+1}\right)+h_j \bm{f}_{j+1}, \\
\text { where } \bm{f}_j &:=\bm{f}\left(\bm{x}\left(t_j\right), \bm{u}\left(t_j\right)\right), \quad h_j:=t_{j+1}-t_j .
\end{aligned}
\label{eq: cubic-lobatto}
\end{equation}

The interpolation function $\tilde{\bm x}$ utilized for $\bm x$ needs to fulfill the continuity at the midpoint of the sample times. The eq.~\ref{eq: cubic-lobatto} satisfies the derivative terms at the boundaries $t_i$ and $t_{i+1}$ are satisfied. The nonlinear program therefore then contains the rest of the collocation constraints in the nonlinear programming problem, which include collocation constraints at the midpoint of $t_i$ and $t_{i+1}$, the inequality constraints at $t_i$, and the boundary conditions at $t_i$ and $t_f$.

\section{Results}
\label{sec:results}

\begin{figure}
    \centering    \includegraphics[width=\linewidth]{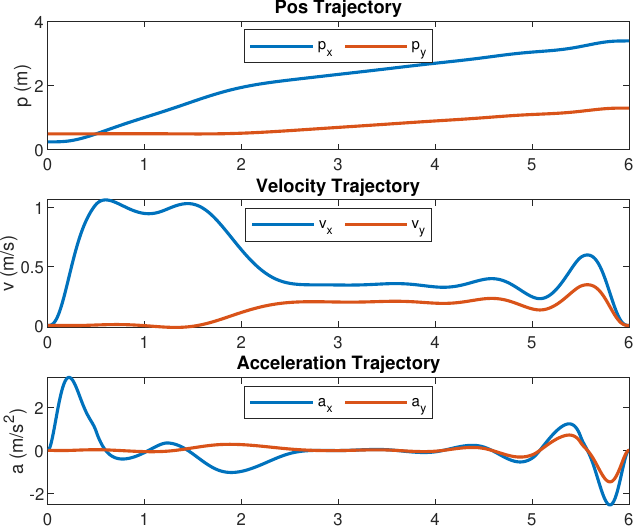}
    \caption{Body states during 30-degree incline WAIR simulation.}
    \label{fig: body states}
\end{figure}

\begin{figure}
    \centering
    \includegraphics[width=\linewidth]{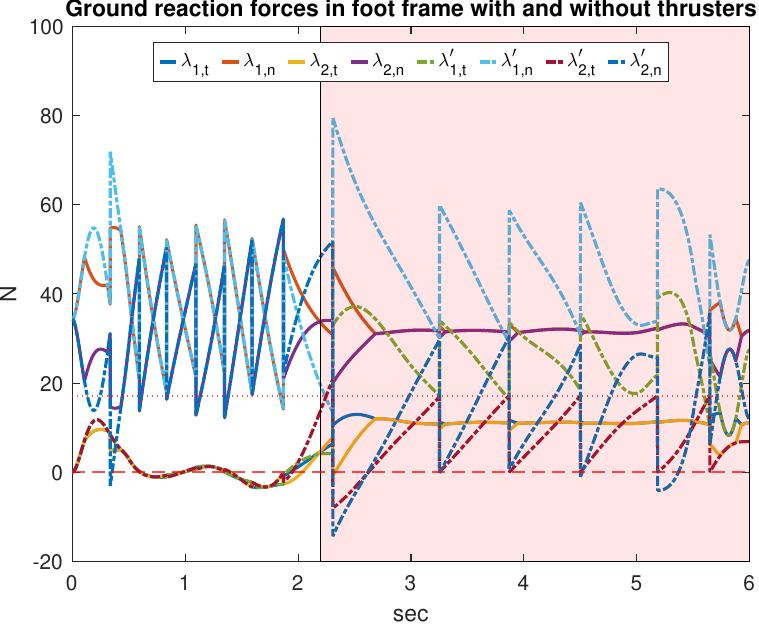}
    \caption{Ground reaction forces of the active stance feet in contact with the ground during the WAIR maneuver with and without thrusters. The red shaded region shows HROM walking on the slope, The horizontal lines at 0 N show that without thruster input the normal forces to the surface tend to become close to zero or even negative (i.e foot loses contact with the ground)}
    \label{fig: GRF}
\end{figure}

\begin{figure}
    \centering
  \includegraphics[width=\linewidth]{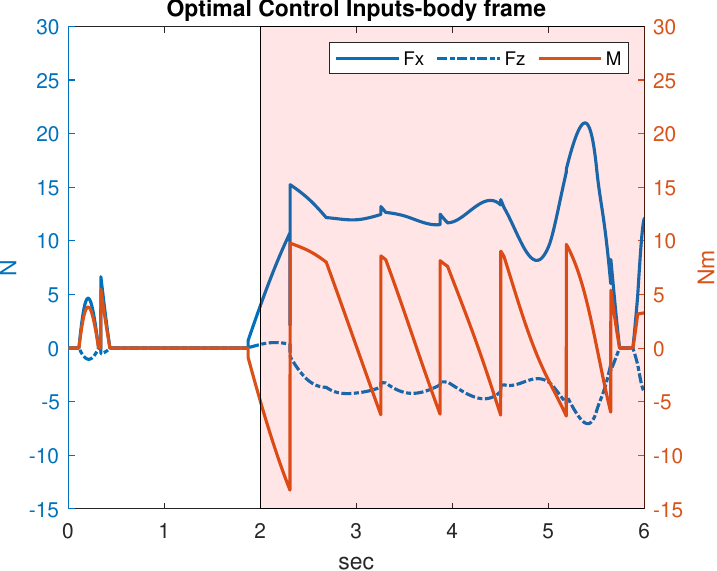}
    \caption{This figure illustrates the optimal thrust wrench inputs in the body frame acting at the COM. The red shaded region shows the robot on the slope.}
    \label{fig: optimal inputs}
\end{figure}

\begin{figure}
    \centering
    \includegraphics[width=\linewidth]{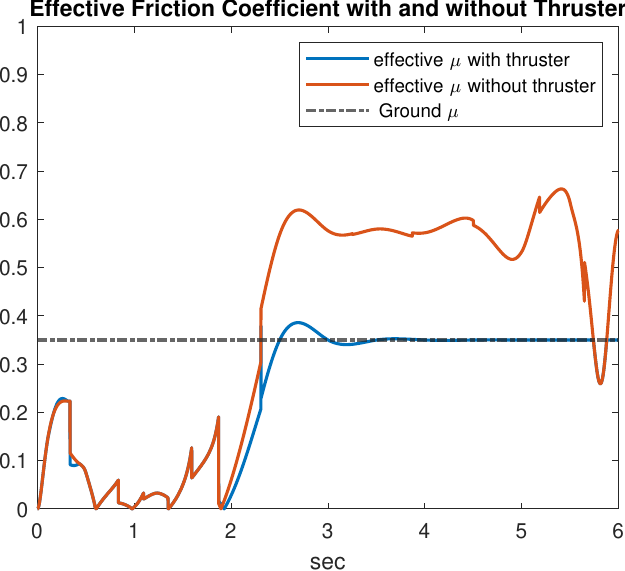}
    \caption{This shows the effective friction coefficient values of the active front stance leg. The black line depicts the friction coefficient of the ground at $\mu$ = 0.35. This shows the effectiveness of the thrusters to regulate the GRF within the friction cone constraints }
    \label{fig: effective friction coeff}
\end{figure}

This simulation was performed in the MATLAB environment using a computer with an Intel core i7 processor and utilized the HROM framework, supported by MATLAB animations, to model and analyze the system's behavior. For the purposes of this simulation, frontal plane movement was constrained. 

A target trajectory for the body was generated, and given a simple heuristic based on the velocity of the body, an active set of footholds are selected from a set of predefined fixed footholds along the trajectory. While generating the target trajectory, the target body pitch at every point is set to be roughly equal to the slope of the ground , i.e, the robot's body is made to be parallel to the ground. Bezier control points were then generated for the swing legs based on some user-defined inputs making it possible to define a target foot location for each leg. The simulation uses a two point contact gait, where diagonally opposite leg pairs synchronized while the remaining pair operated out of phase.

Figure~\ref{fig: snapshots} shows snapshots at different points during the simulation of the WAIR maneuver for a slope of 30 degrees. The WAIR manuever is simulated for the duration of 6 seconds, where the robot walks forward on a flat region for 2 secs and on the slope for 4 secs. While walking on the slope, the body moves forwards along the slope by 1.5 meters. The body trajectories during the simulation are illustrated in Fig~\ref{fig: body states}. 

The optimal control inputs for this sagittal plane constrained simulation then only has 3 inputs which form a 2D wrench. This is plotted in Fig~\ref{fig: optimal inputs}.Interestingly, the optimal thruster inputs show a similar behaviour that is akin to aerodynamic forces generated by the birds in the WAIR maneuver \cite{dial_wing-assisted_2003}. In the flat region, the optimizer does not produce any significant thruster forces, suggesting that it is not necessary for flat ground walking. Whereas, on the slope, the forces are directed in the forward direction and downwards in the body frame, suggesting that the wrench is helping the body to maintain posture on the slope while also allowing the feet to generate sufficient tractive force and maintain a minimum normal force with respect to the surface normal. This can be seen in the active ground reaction force plot in Fig~\ref{fig: GRF}. Without thruster, the normal reaction force tend to become very small and sometimes even negative. With the thruster wrench, the optimizer ensures the forces to be within the friction cone and also ensures that the normal forces are above a specific minimum value. The friction coefficient $\mu $ to simulate the ground is estimated as 0.35 and the optimizer is also able to find the wrench that respects the corresponding friction cone constraints (See friction cone in snapshots Fig~\ref{fig: snapshots}, and Fig~\ref{fig: effective friction coeff}).

\section{Concluding remarks}
\label{sec:conclusion}
Inspired by the WAIR maneuver seen in galliform birds, we employed a MATLAB simulation of Husky robot using HROM to perform the WAIR maneuver. The WAIR maneuver requires co-ordination of the joints and the thrusters. In our research, we use a polynomial approximation of the reduced order dynamics of Husky, called the HROM. We see that the controller is able to find the optimal thruster wrench and regulate the ground reaction forces. This observation also provides an insight into the different ways we can exploit the thruster wrench for any a large variety of terrains. 

By exploiting the unique design of Husky, our future research pathways would be to implement the proposed optimal controller on the physical robot. Further collaboration between the legs and the thruster inputs can be achieved if accurate torque control of the joints are achieved. This can lead to a more sophisticated WAIR maneuver with the legs also actively contributing in regulating the ground forces. 
A hybrid impact-model, similar to ones more commonly used in bipedal locomotion, would also help to more accurately estimate the ground reaction forces at the end of every swing phase. Combined with the above, a focused effort to design an optimal controller effective against disturbances and noise as commonly expected in physical systems needs to be carried out.

\printbibliography
\end{document}